\pdfoutput=1

\documentclass[11pt]{article}

\usepackage{acl}

\usepackage{times}
\usepackage{babel}
\usepackage{amssymb}
\usepackage{footmisc}
\usepackage{makecell}
\usepackage{arydshln}
\usepackage{latexsym}
\usepackage{hyperref}
\usepackage{url}
\usepackage{paralist}
\usepackage{pifont}
\usepackage{booktabs}
\usepackage{listings}

\usepackage[T1]{fontenc}

\usepackage[utf8]{inputenc}
\usepackage{graphicx} 

\usepackage{microtype}
\usepackage{float}
\usepackage{graphicx}
\graphicspath{{figures/}}
\usepackage{listings}
\usepackage{xcolor}
\usepackage{lipsum}
\usepackage{hyperref}

\definecolor{codegreen}{rgb}{0,0.6,0}
\definecolor{codegray}{rgb}{0.5,0.5,0.5}
\definecolor{codepurple}{rgb}{0.58,0,0.82}
\definecolor{backcolour}{rgb}{0.95,0.95,0.92}

\lstdefinestyle{mystyle}{
    backgroundcolor=\color{backcolour},   
    commentstyle=\color{codegreen},
    keywordstyle=\color{magenta},
    numberstyle=\tiny\color{codegray},
    numbers=none,
    stringstyle=\color{codepurple},
    basicstyle=\ttfamily\scriptsize,
    breakatwhitespace=false,         
    breaklines=false,                 
    captionpos=b,                    
    keepspaces=true,                 
    numbers=left,                    
    numbersep=5pt,                  
    showspaces=false,                
    showstringspaces=false,
    showtabs=false,                  
    tabsize=2
}
\definecolor{lightergray}{RGB}{230,230,230}
\definecolor{DarkGreen}{RGB}{30,130,30}
\newcommand{\mimir}{\raisebox{-5.5pt}{\includegraphics[scale=0.025]{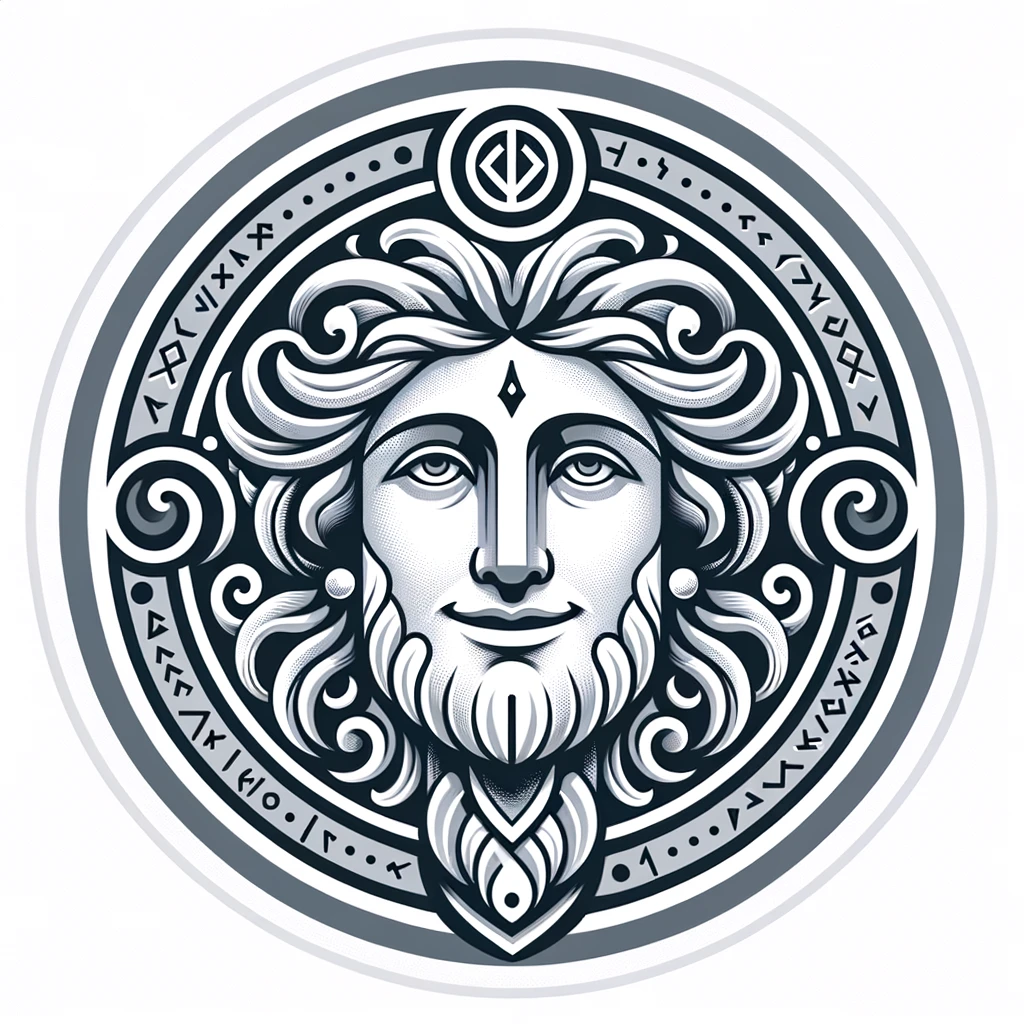}}}
\newcommand{\cmark}{\textcolor{DarkGreen}{\ding{51}}}
\newcommand{\xmark}{\textcolor{red}{\ding{55}}}%

\title{ \mimir{}  \textsc{MIMIR}: A Streamlined Platform for Personalized Agent Tuning \\ in Domain Expertise}

\author{Chunyuan Deng{$^{1}$}, Xiangru Tang{$^{2}$, Yilun Zhao{$^2$}, \textbf{Hanmin Wang}{$^{2}$}\textbf{,} \textbf{Haoran Wang}{$^{2}$}\textbf{,} }\\\textbf{Wangchunshu Zhou}{$^3$}\textbf{,} \textbf{Jiannan Cao}{$^4$}\textbf{,} \textbf{Arman Cohan}{$^2$}\textbf{,} \textbf{Mark Gerstein}{$^2$} \\
{$^1$} Georgia Institute of Technology {$^2$} Yale University {$^3$} ETH Zurich {$^4$} MIT  \\
{\tt{{cdeng73@gatech.edu}}}}

\begin{document}
\maketitle
\begin{abstract}
Recently, large language models (LLMs) have evolved into interactive agents, proficient in planning, tool use, and task execution across a wide variety of tasks. However, without specific agent tuning, open-source models like LLaMA currently struggle to match the efficiency of GPT- 4, particularly given the scarcity of agent-tuning datasets for fine-tuning. In response, we introduce \textsc{Mimir}: a streamlined platform offering a customizable pipeline that enables users to leverage both private knowledge and publicly available, legally compliant datasets at scale for \textbf{personalized agent tuning}. Additionally, \textsc{Mimir} supports the generation of general instruction-tuning datasets from the same input. This dual capability ensures that language agents developed through the platform possess both specific agent abilities and general competencies. \textsc{Mimir} integrates these features into a cohesive end-to-end platform, facilitating everything from the uploading of personalized files to one-click agent fine-tuning. 
\end{abstract}

\section{Introduction}

Recently, large language models (LLMs) have undergone a significant evolution, transitioning into interactive agents that have demonstrated considerable progress in a multitude of scenarios~\cite{chatgpt,GPT4,claude,bard}. The commendable performance of these models across various downstream tasks has incited researchers to propose methods for utilizing LLMs to generate instruction datasets~\cite{peng2023instruction,wang2023selfinstruct,sun2023principledriven}. The quality and diversity of such data are instrumental in aligning, pre-training, and fine-tuning LLMs~\cite{sun2023principle, chiang2023vicuna,taori2023stanford,xu2023baize,li2023camel,shao2023synthetic,ding2023enhancing}. Besides promoting methods for general instruction tuning to enhance the capability of LLMs, there is increasing research emphasis on fine-tuning LLMs to acquire tool usage~\cite{schick2023toolformer,Zhang2024CodeAgentEC,zhou2023agents} and establish stronger agent abilities~\cite{Chen2023FireActTL,qin2023toolllm,zeng2023agenttuning}.

However, in contrast to the abundant datasets available for instruction tuning~\cite{wang2023selfinstruct,chiang2023vicuna,xu2023baize,li2023camel,ivison2023camels}, datasets focused on agent tuning~\cite{zeng2023agenttuning} are in short supply. This imbalance has inculcated reliance on proprietary LLMs such as ChatGPT or GPT-4 as mainstay tools for cognitive processing and planning. This dependency propels us into a sphere of uncertainty, especially when integrating confidential domain knowledge into model training as highlighted by \citet{Kim2023ProPILEPP} and \citet{tian2023evil}. Further, concentrating solely on fine-tuning LLMs with tool-learning datasets might inadvertently affect their broad-spectrum capabilities, a concern raised by \citet{zeng2023agenttuning}.
For example, An LLM, trained on a dataset focused on diagnosing cardiovascular diseases, could potentially miss a crucial anomaly that falls outside its training data, such as an unusual symptom of a rare disease. Additionally, the process of incorporating confidential patient data into a proprietary model can also raise serious privacy concerns.

\begin{figure*}[htbp]
\centering
\includegraphics[width=1.0\textwidth]{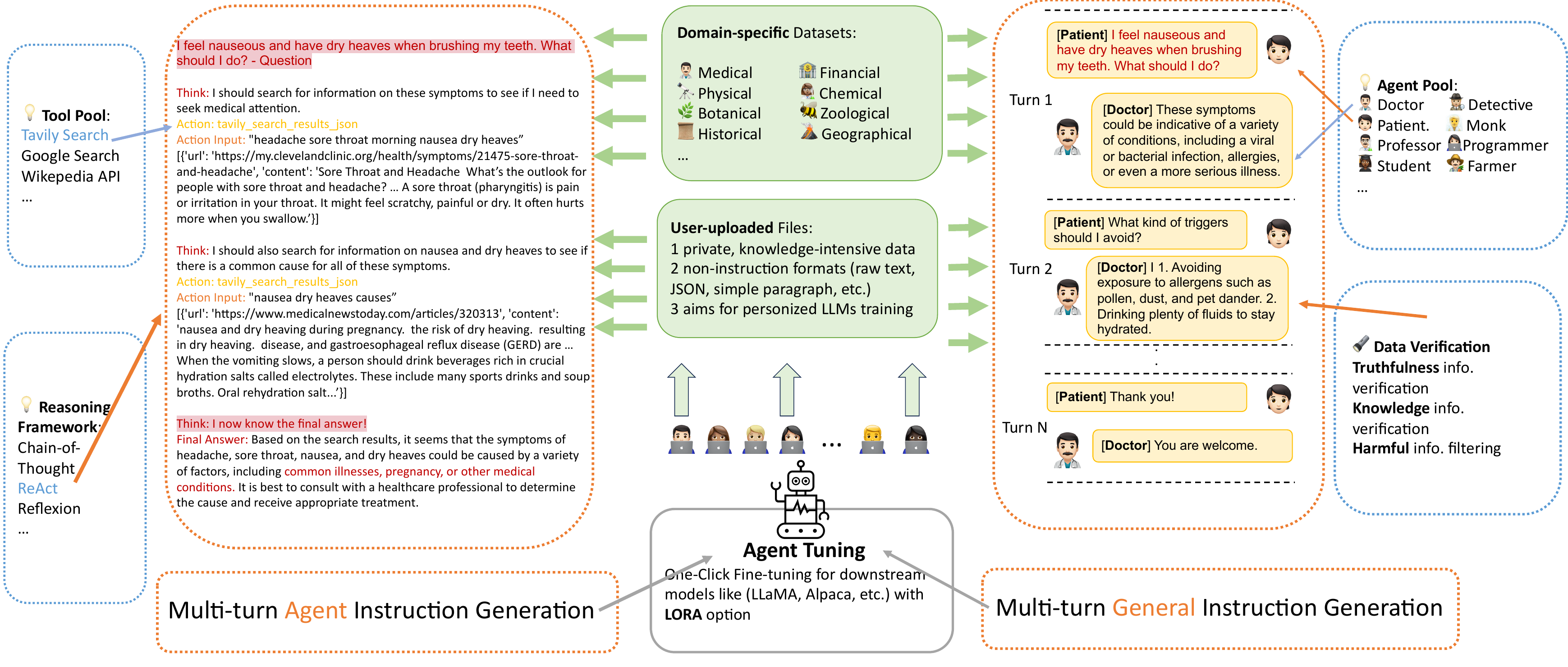}
\caption{\textit{MIMIR} provides an integrated pipeline from generating \textbf{multi-role} and \textbf{multi-turn} instructions to a one-click fine-tuning process for downstream models. Users also could upload their files combined with existing domain-specific datasets to customize their instruction data.}
\label{fig:flowchart}
    \vspace{-.3cm}
\end{figure*}

In this paper, we introduce MIMIR\footnotemark[1], a novel streamlined platform, as illustrated in Figure \ref{fig:flowchart}. MIMIR is adept at tackling the challenges prevalent in specialized fields such as medicine, law, and finance. 
A primary obstacle in these domains is the handling of sensitive, non-standard data formats abundant with sector-specific knowledge.
\textsc{MIMIR} facilitates the integration of proprietary knowledge data with an established, external domain-specific knowledge base. Through this integration,  \textsc{MIMIR} simultaneously generates a multi-turn agent tuning dataset. 
For the creation of the general multi-turn instruction tuning dataset, we adopt the method proposed in \citet{park2023generative}, employing LLMs as interactive agents in multi-round conversations. Regarding the agent tuning dataset,  \textsc{MIMIR} integrates reasoning frameworks and search tools to generate interaction trajectories. This allows users to tailor templates within these frameworks, offering demonstrations that align with their specific objectives.
\footnotetext[1]{The code is avaialable at \href{https://github.com/Miraclemin/mimir}{MIMIR}.}
Our pipeline seamlessly incorporates private and public knowledge bases, agent-tuning data generation protocols, multi-role agent configurations, and one-click fine-tuning into a unified flow. The multi-agent-based data generated through this pipeline has higher accuracy and credibility.
Notably, in comparison to original data, self-instruct~\cite{wang2023selfinstruct}, and Baize~\cite{xu2023baize}, we achieved win or equal rates of 87\%, 75\% and 77\%, respectively.  We summarize the key features of  \textsc{MIMIR} as follows:
\begin{compactitem}
\item {\bf Simple and User-friendly.}
For users who are unfamiliar with agent tuning, it is feasible to activate agent capabilities using open-source models such as LLaMA, facilitating the creation of personalized agents.
\item {\bf Private and Dataset Integration.}
Users can seamlessly integrate public datasets with their proprietary knowledge bases using  \textsc{MIMIR} offline, thereby ensuring data privacy and avoiding leakage issues.
\item {\bf Domain-Specific Role-playing.}
Our system supports domain-specific role-playing during the generation of domain-specific data. Notably, it facilitates multi-turn interactions among various medical roles, including doctors, patients, and medical students, for the creation of medical domain data.
\item {\bf One-Click Fine-Tuning.}
Using parameter visualization and LoRA technology~\cite{hu2021lora}, users can formulate and implement customized fine-tuning scripts for LLMs, thereby optimizing performance and efficiency.
\end{compactitem}

\vspace{-.2cm}
        \vspace{-.1cm}

\section{Background and Related Work} \label{sec:related-work}

        \vspace{-.1cm}

\paragraph{Domain-Specific Instruction Data Generation}
Following the success of ChatGPT \cite{chatgpt} and GPT-4 \cite{GPT4}, open-source LLMs like LLaMA \cite{touvron2023llama}, Alpaca \cite{taori2023stanford} and Mistral \cite{jiang2023mistral} have arisen, all requiring instruction data for training. Although these LLMs exhibit remarkable performance in general domains, their lack of domain-specific knowledge results in inadequate performance in areas that require specialized expertise. Several efforts have been made to adapt LLMs to these domain-specific areas. This typically involves generating domain-specific data to fine-tune such LLMs, as seen in examples like HuaTuo \cite{wang2023huatuo}. LLMs fine-tuned on domain-specific datasets have shown significant performance improvements in their respective domains.

\begin{table*}[t]
    \centering
    \resizebox{0.97\linewidth}{!}{
    \begin{tabular}{lccccc}
        \toprule
        \textbf{Resource} & \makecell{\textbf{MIMIR}\\(ours)}& \makecell{ \textbf{Self-Instruct} \\\citep{wang2023selfinstruct} } & 
        \makecell{ \textbf{Baize}\\\citep{xu2023baize} } &  \makecell{ \textbf{AgentInstruct} \\\citep{zeng2023agenttuning} } & \makecell{ \textbf{FireAct}\\\citep{Chen2023FireActTL} } 
        \\
         \cmidrule(lr){1-1}  \cmidrule(lr){2-2}  \cmidrule(lr){3-3}  \cmidrule(lr){4-4}  \cmidrule(lr){5-5} \cmidrule(lr){6-6}
         Real API Call? & \cmark & \cmark & \cmark & \cmark & \cmark  \\ 
          Multi-step Reasoning?   & \cmark & \xmark & \cmark & \cmark & \cmark  \\ 
          API Retrieval?   & \cmark & \xmark & \xmark & \cmark & \cmark  \\ 
          
          Instruction Tuning for Tool Learning?   & \cmark & \xmark & \xmark & \cmark & \cmark  \\ 
          Instruction Tuning for Alignment?   & \cmark & \cmark & \cmark & \cmark & \xmark  \\ 
          Role-playing for Generation?   & \cmark & \xmark & \xmark & \xmark & \xmark  \\ 
          \noalign{\vskip 0.5ex}\hdashline\noalign{\vskip 0.5ex}
          Expertise Focus   & Domain-specific & General & General & Task-specific
          & Task-specific  \\ 
          Domain and Tasks   & Medical, Law, Science, ... & General & Chat & Web, KG, OS, Database & Question Answering  \\ 
          \noalign{\vskip 0.5ex}\hdashline\noalign{\vskip 0.5ex} 
          Number of Datasets   & $\mathbf{520}$ & - & - & $6$ & $4$ \\ 
          Avg. Reasoning Traces  & Customized & 1.0 & Customized & $5.24$ & Customized \\ 
         \bottomrule
    \end{tabular}
    }
    \caption{A comparison of our  \textsc{MIMIR} to other instruction tuning datasets for tool use and general ability. KG and OS stand for knowledge graph and operation systems.}
    \label{tab:dataset_comparison}
        \vspace{-.3cm}
\end{table*}

\paragraph{Agent Tuning}
Recently, LLMs excel in text understanding and following instructions \cite{qian2023communicative, chiang2023can, shen2023large, gao2023human}. Beyond its single-agent capabilities, agents further allow for the customization of multi-agent systems. Such systems are valuable in specific domains like medical domain~\cite{tang2023medagents}, games \cite{park2023generative}, software development \cite{qian2023communicative}, and data generation \cite{xu2023baize, li2023camel}. Research suggests that through mechanisms such as debate and cooperation, the collective capabilities of agents can not only be enhanced but also lead to the improvement in the quality of generated responses \cite{li2023camel, liang2023encouraging}. As a result, there is increasingly more work utilizing multi-agent systems for data generation \cite{du2023improving, li2023camel, qian2023communicative}. In a Multi-Agent based data generation system, individual LLM agents can generate instruction data through dialogue. Besides utilizing the agent's ability to generate instruction tuning data, there are also some methods to generate data for agent tuning~\cite{zeng2023agenttuning,Chen2023FireActTL}, which focus more on tasks like web using and reasoning or computer controlling. 

\section{System Design and Workflow}

\subsection{System Input}

\paragraph{Self-defined topics}
In offline settings, users often hold private domain-specific knowledge data. Recognizing this,  \textsc{MIMIR} features an offline pipeline that allows users to import their sector-specific knowledge. This approach is designed not only to safeguard privacy but also meet distinct user requirements. Considering the prevalence of domain-specific data among users,  \textsc{MIMIR} is adeptly designed to accommodate custom inputs from the user's side. As shown in Table~\ref{tab:input_example},  \textsc{MIMIR} accepts two types of inputs for users to upload their files.

\begin{table}[t]
    \centering
    \small
    \begin{tabular}{p{1.5cm}|p{5cm}}
    \toprule
         Topic Type& Examples of User-defined Input\\
    \midrule
         Keyword-based & ``Anatomy'', ``Biochemistry'', ``Biostatistics'', ``Cardiology'', ``Dermatology'', ``Emergency Medicine'', ``Endocrinology'', ``Epidemiology'', ``Gastroenterology''. \\
    \midrule
         Sentence-based& ``In ophthalmology, cataracts, which are characterized by the clouding of the eye's natural lens, are a leading cause of visual impairment worldwide and can be effectively treated through a surgical procedure that replaces the clouded lens with an artificial one.''\\
    \bottomrule
    \end{tabular}
    \caption{User-Defined topic examples: Users have the option to upload their private domain-specific knowledge by two types of input: keyword or sentence.}
    \label{tab:input_example}
        \vspace{-.3cm}

\end{table}

\paragraph{Domain-specific Dataset Incoporation}
In addition to leveraging parametric knowledge in Large Language Models,  \textsc{MIMIR} enhances its capabilities by incorporating 520 domain-specific datasets available on the Internet. This integration serves as a robust supplementary knowledge base for instructional data. For instance, in the medical domain,  \textsc{MIMIR} includes several public medical datasets similar to the setting in Flan-PaLM~\cite{singhal2023expertlevel} into our pipeline: MedQA~\cite{jin2020disease}, MedMCQA~\cite{pmlr-v174-pal22a}, PubMedQA~\cite{jin-etal-2019-pubmedqa}, MMLU Clinical Topics~\cite{hendrycks2021measuring}. As Shown in Table~\ref{tab:dataset_comparison},  \textsc{MIMIR} integrates 520 datasets, utilizing a reasoning framework and retrieval tools to generate user-specific trajectory interactions.

\subsection{Agent Tuning Data Generation}

\subsubsection{Multi-turn General Instruction Data}

\paragraph{Multi-turn Dialogue}
After users select their self-uploaded topics and an existing domain-specific dataset for generating the instruction dataset,  \textsc{MIMIR} seamlessly integrates these datasets in the backend to create an intermediate data pool. Each data point in this pool is utilized as a keyword or key sentence in the subsequent step. Building on previous work~\cite{xu2023baize}, we generate a multi-turn dataset based on multiple rounds of interaction between a human and an agent. Additionally, we provide the functionality for users to predefine the number of interaction rounds they wish to include in their instruction data, ensuring tailored dataset generation. Compared to Camel~\cite{li2023camel}, which employs role-playing and inception prompting for agent communication, our method focuses more on generating diverse, domain-specific instruction datasets rather than solving reasoning tasks. In our agent setting, we do not use complex communication to interact with environments. Instead, we employ a simpler inception prompting approach, enabling LLMs to role-play agents and generate representative data (e.g., role-playing as doctors, medical professors, and students) to create medical instruction data in the medical domain.

\paragraph{Domain-specific Role-playing}
We leverage LLMs to replicate specific domain roles via advanced inception prompting. Within the medical sphere, our system comprises 14 unique roles: Doctor, Nurse, Pharmacist, Medical Laboratory Technician, Physical Therapist, Nutritionist, Psychologist, Radiology Technician, Medical Researcher, Medical Educator, Medical Administrator, Medical Interpreter, Medical Equipment Engineer, and Medical Librarian. Our methodology for role-specific prompting is both simple and efficient, particularly adept at producing multi-turn instructional data. We have meticulously crafted a bespoke prompt setting for each role in our  \textsc{MIMIR} agent ensemble. This allows users to select the most relevant in-domain roles for generating multi-turn instruction tuning datasets. This approach significantly surpasses the efficacy of previous default configurations.

\subsubsection{Multi-turn Agent Instruction Data}

\paragraph{Initial Trajectory}
For datasets specifically tailored for \textsc{MIMIR}, we primarily utilize their training split as our input source. In cases where datasets are not partitioned, we employ the entire dataset for training purposes. The training set examples are used directly as the initial trajectory. For a limited number of datasets that do not follow an instruction-based format, we leverage GPT-4 for synthesizing the initial trajectory. For instance, a phrase like ``headaches, sore throat, dry heaves'' is transformed into a more contextualized statement: ``Recently, I’ve been experiencing headaches and a sore throat. In the mornings, I feel nauseous, especially when brushing my teeth, accompanied by dry heaves. What should I do?''.

\vspace{-.1cm}
\paragraph{Tool}
We have augmented \textsc{MIMIR} with a suite of search tools. Following the approaches laid out in~\citet{press2023measuring} and~\citet{Chen2023FireActTL}, we integrated SerpAPI\textsuperscript{1}\footnote{\textsuperscript{1} \url{https://serpapi.com/}} to develop a Google search tool. This tool is designed to retrieve the foremost relevant item, prioritizing data from the ``highlight words''. Additionally, \textsc{MIMIR} is equipped with Tavily\textsuperscript{2}\footnote{ \textsuperscript{2} \url{https://tavily.com/}} as an alternative search API. These search tools empower LLMs with the latest knowledge and information pertinent to their development trajectory, thereby facilitating robust agent tuning. This integration is crucial for ensuring that LMs remain up-to-date and effective in their responses.

\vspace{-.1cm}

\paragraph{Reasoning Framework}
Within our \textsc{MIMIR} framework, we incorporate ReAct~\cite{yao2023react} as our primary reasoning framework to generate rationales. For each interaction cycle, this framework outputs two components: the thought process, which reflects on previous results, and the action, which involves selecting and utilizing tools. As an example, it might use the Google Search tool to acquire necessary information. Following this, the action yields a result, such as search outcomes, within our framework. If the thought process aligns with the correct direction, it concludes in the thinking phase, leading to the final answer. Besides the default reasoning framework,  \textsc{MIMIR} also supports user-customized Chain-of-Thought (CoT)~\cite{wei2023chainofthought} Templates and Reflexion~\cite{shinn2023reflexion} mechanisms. These additional mechanisms cater to varying user preferences and contribute to the system's versatility. Importantly, the decision-making steps within  \textsc{MIMIR} are directed by the internal reasoning processes provided by these frameworks. This design ensures a coherent and efficient reasoning pathway, tailored to each specific interaction.

\begin{figure}[ht]
    \centering
    \includegraphics[width=0.51\textwidth]{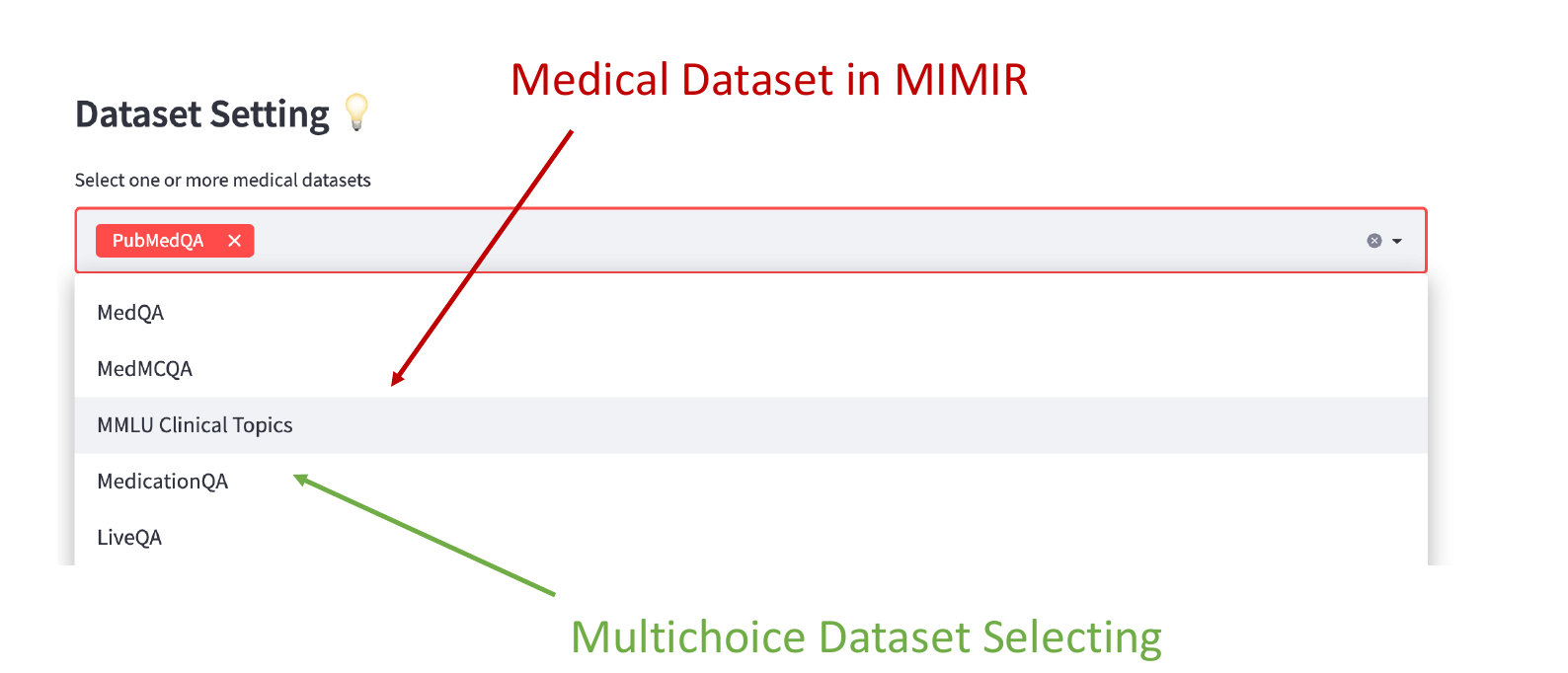} 
    \caption{User can upload the custom data and select multiple domain-specific datasets in MIMIR. In this figure, we provide an example for selecting medical domain datasets.} %
    \label{fig:data_selection}
        \vspace{-.4cm}
\end{figure}

\begin{figure}[t]
    \centering
    \includegraphics[width=0.535\textwidth]{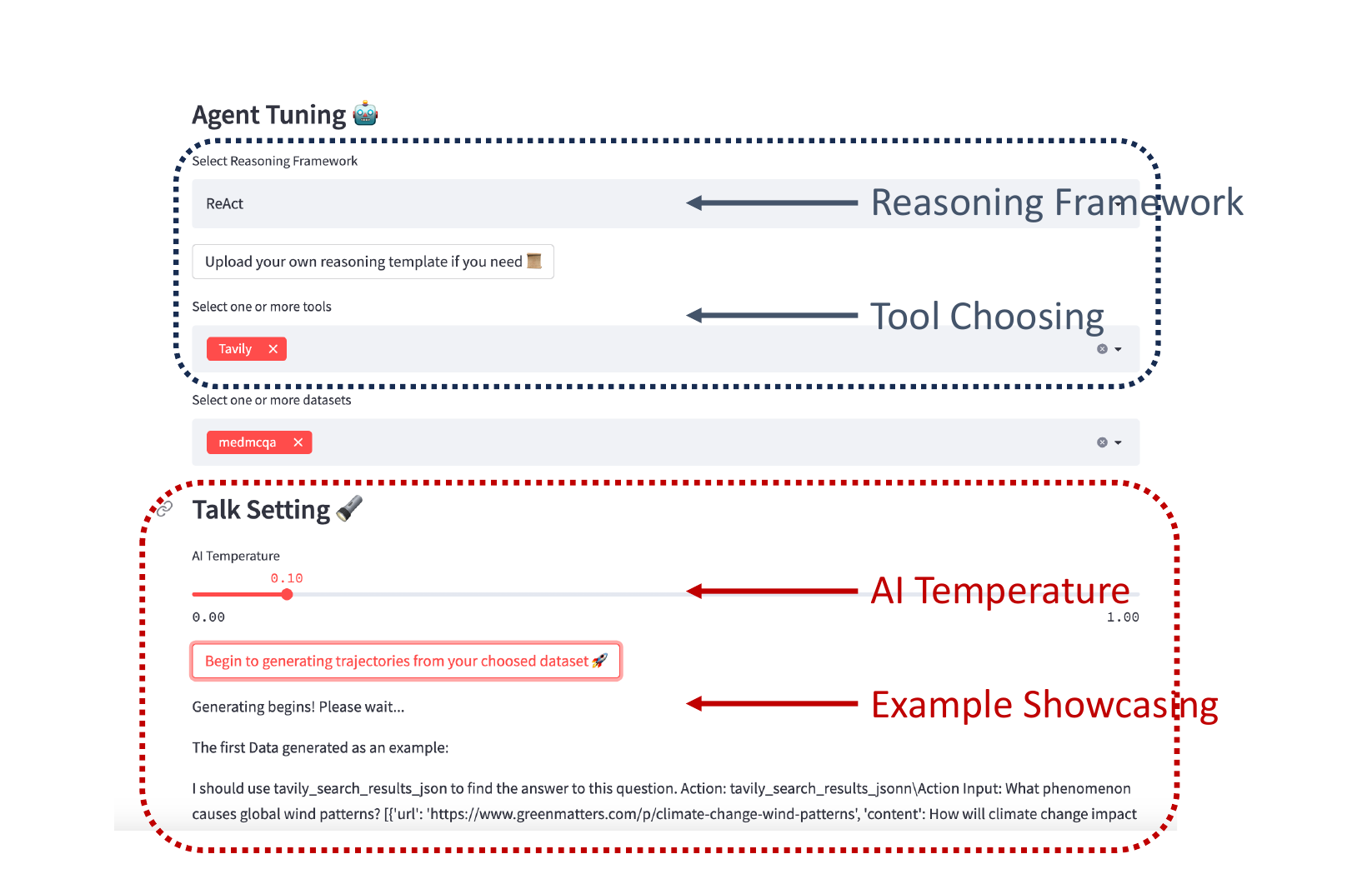} 
    \caption{Agent tuning interface: we commence by allowing the user to select a reasoning framework and designate their preferred tools. Subsequently, we empower the user to configure the hyperparameters for the models. Furthermore, to facilitate a comprehensive understanding, we provide an illustrative example as the user proceeds with dataset generation.} 
    \label{fig:agent_tuning}
        \vspace{-.2cm}
\end{figure}

\begin{figure}[t]
    \centering
    \includegraphics[width=0.55\textwidth]{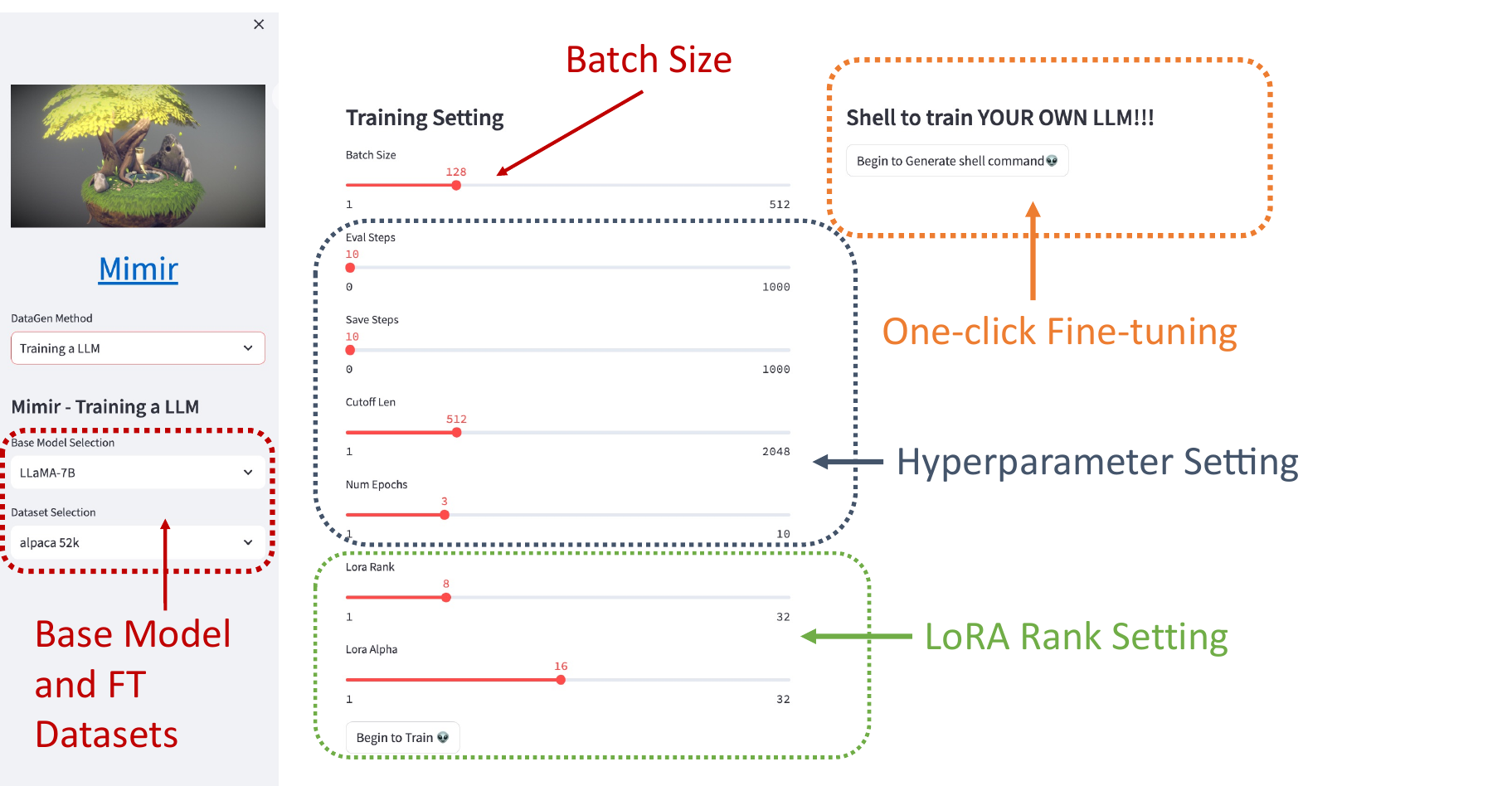} 
    \caption{User can fine-tune personalized models with a single click, selecting from a pre-defined list of models such as LLaMA.} 
    \label{fig:train}
    \vspace{-.2cm}
\end{figure}

\section{The \textit{MIMIR} UI}

The system design of our framework focuses on enabling users to create instructional data to enhance the capabilities of Large Language Models (LLMs). In this section, we present three interface screenshots (S1, S2, and S3), accompanied by detailed instructions, to demonstrate the design of the MIMIR UI. 

\paragraph{S1: Dataset Selection View}
As depicted in Figure~\ref{fig:data_selection}, MIMIR facilitates the selection of multiple domain-specific datasets through an efficient and user-friendly multi-selection checkbox interface. Given the extensive collection of 520 datasets, users can conveniently search by entering dataset initials in the provided search box. This feature allows users to efficiently narrow down their options and locate the most relevant datasets for their needs.
\paragraph{S2: Agent Tuning View}
In Figure~\ref{fig:agent_tuning}, we present MIMIR, a comprehensive platform that amalgamates various reasoning frameworks and tools to facilitate the generation of rationales for trajectory interactions. Initially, users select a reasoning framework from options including CoT~\cite{wei2023chainofthought}, ReAct~\cite{yao2023react} (the default choice), and Reflexion~\cite{shinn2023reflexion}. Subsequently, they can choose from a suite of tools available in our tool pool. Additionally, users have the flexibility to upload custom templates to create CoT rationales tailored to their specific requirements. The next step involves selecting and uploading multiple datasets as source input. Finally, users can create agent-tuning datasets by clicking the designated button at the bottom of the interface. To enhance user experience and ensure the quality of the generated datasets, we also provide an example showcasing the dataset generation process.

\paragraph{S3: One-Click Finetuning}
The training script interface, as depicted in Figure~\ref{fig:train}, enables users to fine-tune foundation models such as LLaMA with a single click, using datasets they have created. This can be done in either our default or LoRA~\cite{hu2021lora} settings. Furthermore, the interface provides the functionality to create data scripts for model fine-tuning, leveraging visualized parameters. This innovative feature empowers users to efficiently train large-scale models tailored to their specific domains, utilizing the dialogue data they have generated. 

\section{Community Guidelines}
The objective of these guidelines is to establish a uniform framework for the development, validation, and application of agent-tuning instructions within the MIMIR system, with a specific focus on addressing the unique demands and challenges associated with domain expertise.

We recognize that the dynamic integration of public datasets into the system presents unique challenges, particularly regarding copyright and appropriate use. This implies that if a dataset encounters copyright issues or is unexpectedly removed, we must also withdraw it through an automated process. In practical terms, this means we often face various issues daily due to these constraints. Therefore, as a community, we must adhere to these guidelines within the legal boundaries to ensure compliance and maintain the integrity of our system.

Continuing from the established guidelines, it is crucial to emphasize the importance of ethical considerations and data privacy in the handling of datasets. As we navigate the complexities of incorporating publicly sourced data, we must remain vigilant in protecting the privacy and rights of individuals represented within these datasets. In this way, we can ensure that our pursuit of technological advancement and domain expertise does not come at the expense of ethical responsibility and user trust.
\label{sec:contrib}

\section{Human Evaluation of the Generated Data}

\paragraph{Domains}
In our study, we carefully selected a diverse set of data samples to form our investigation set for source input in MIMIR. Specifically, we chose 25 random samples from each of the following medical domain datasets: MedQA~\cite{jin2020disease}, MedMCQA~\cite{pmlr-v174-pal22a}, PubMedQA~\cite{jin-etal-2019-pubmedqa}, MMLU Clinical Topics~\cite{hendrycks2021measuring}. Our approach ensures a comprehensive dataset, facilitating an in-depth analysis of medical domain data.

\begin{figure}[ht]
\begin{center}
        \centering
        \includegraphics[scale = 0.38]{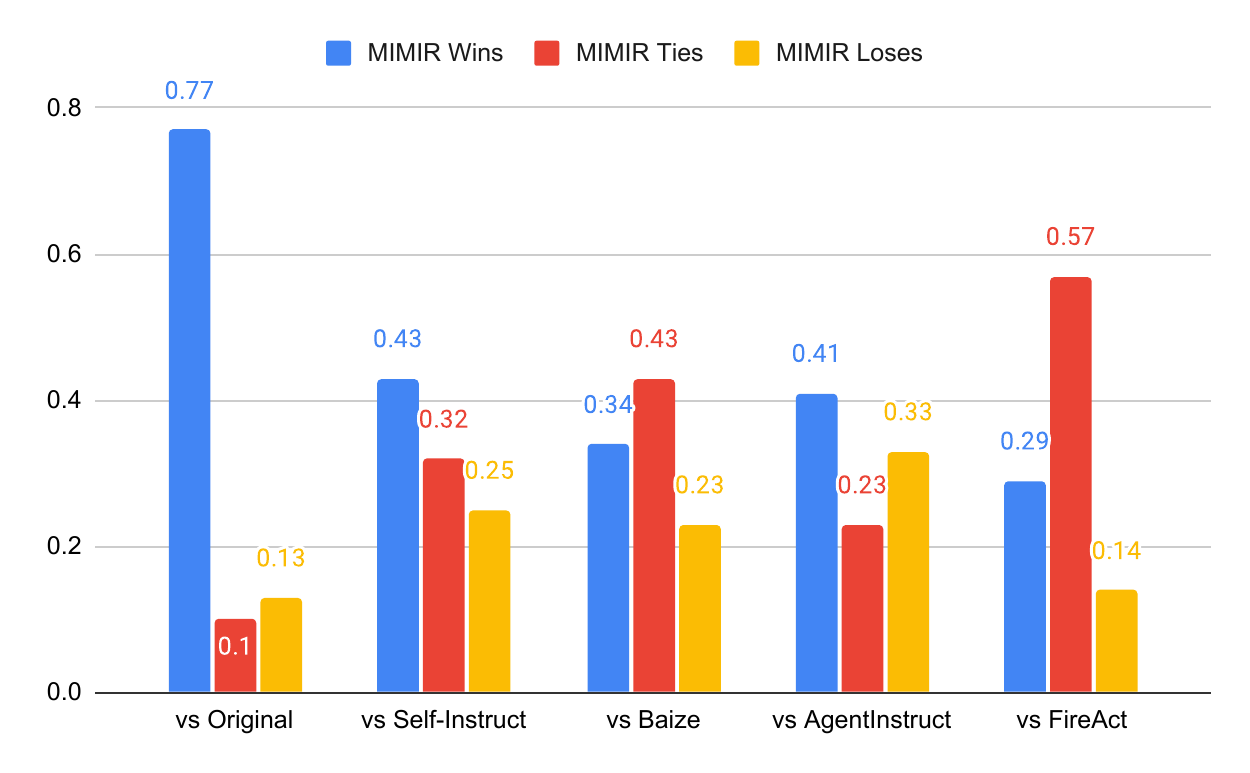}
    \caption{General Preference of data generated by \textsc{MIMIR} in 3-turn setting with \emph{original topics}, \emph{self-instruct}~\cite{wang2023selfinstruct}, \emph{Baize}~\cite{xu2023baize}, \emph{AgentInstruct}~\cite{zeng2023agenttuning} and \emph{FireAct}~\cite{Chen2023FireActTL}.}
        \label{fig:alignment}
    \hfill
\end{center}
\vspace{-0.6cm}
\end{figure}

\paragraph{Experiment Setting}
We utilized the default configuration in \textsc{MIMIR} to process the input source datasets for generating instruction data. MIMIR's token limit is set at 1000, with a temperature setting of 0.1. We conducted a comparative analysis of \textsc{MIMIR} with Baize~\cite{xu2023baize}, Self-Instruct~\cite{wang2023selfinstruct}, AgentInstruct~\cite{zeng2023agenttuning} and FireAct~\cite{Chen2023FireActTL} using identical settings to produce output responses. Given the high costs associated with human evaluation, we engaged 13 domain experts to select the most appropriate output from these four methodologies. These experts were instructed to complete the evaluation sets independently, relying solely on their professional judgment, without any intercommunication. The findings of this study are presented in Figure~\ref{fig:alignment}.

\paragraph{Result and Observation}
Our findings reveal that, in comparison to original datasets, \textsc{MIMIR} shows a marked preference. The reasoning behind this trend is that outputs from \textsc{MIMIR} were chosen more frequently by domain experts than those yielded from simpler topics. When set against methods that primarily follow instructions, like self-instruct and Baize, \textsc{MIMIR} displays significant strides by demonstrating enhanced capabilities when enabling agent learning. Moreover, \textsc{MIMIR} showcases considerable potential in assimilating external, domain-specific knowledge, particularly when compared with other agent tuning frameworks such as FireAct and AgentInstruct.

\section{Conclusion and Future Work}

MIMIR is a streamlined platform designed for personalized language agent tuning, focusing on Domain Expertise. It integrates domain-specific datasets along with user-uploaded topics, utilizing various contemporary reasoning frameworks and tools. This integration facilitates the generation of both general and specific tuning datasets for personalized agent training. For future development, enhancing  \textsc{MIMIR} by incorporating diverse tools and automating tool selection appears promising. Additionally, we plan to conduct quantitative evaluations of our generated data, aiming to compare its performance against existing methods.
\section{Ethics Statement}
This paper introduces a streamlined platform for personalized agent tuning, aiming to empower users to refine their agents while ensuring the privacy of their data. 

\paragraph{Privacy} Excluding personal data, all datasets integrated into \textsc{Mimir} are accompanied by licenses that authorize us to compile, adapt, and redistribute the original datasets. Additionally, we introduce a knowledge filtering method to eliminate potentially harmful and inaccurate information. The model and reasoning framework employed do not reveal sensitive information. 

\paragraph{Data} During interactions with human participants, we strictly adhered to ethical standards and prioritized their well-being. The datasets and output examples provided for selection are exclusively sourced from publicly available and legally compliant materials.

\paragraph{Recruitment of Domain Experts}
The 13 domain experts, engaged in our study, were recruited from a variety of sources. An open call for participation was announced in various professional medical forums, online groups, and mailing lists, directly targeting professionals in the medical community. The recruitment process ensured that all potential evaluators possessed relevant qualifications and expertise in the subject matter. Prior to participation, all participants provided informed consent and received a comprehensive briefing about the study's purpose, their expected role, and the data handling procedures to ensure anonymity and data privacy.

\paragraph{Processing of Evaluation Results}
To maintain the ethical standards of word anonymity and confidentiality, all obtained evaluations were anonymized and de-identified before analysis. Evaluators' identities were replaced with arbitrary numerical identifiers to protect their identities during the analysis and subsequent publication processes. Moreover, all data underwent privacy-preserving protocols, and secure, encrypted databases were used for storage to prevent unauthorized access and ensure data integrity.

\section{Limitation}

There are areas in which the system can still improve.

\paragraph{Agent-Tuning Data Generation Protocols:} While \textsc{MIMIR} incorporates private and public knowledge bases and employs multi-role agent configurations, the underlying assumptions could be limiting. The generation protocols might not account for dynamic changes in real-world data or rapid advancements in the knowledge base of specific domains.

\paragraph{Dependence on External Tools:} The efficiency of \textsc{MIMIR} heavily depends on the performance of external tools such as SerpAPI and Tavily. Any limitation inherent in these tools will directly impact the accuracy and results procured by \textsc{MIMIR}.

\paragraph{Domain-Specific Data:} As the effectiveness of \textsc{MIMIR} is closely tied to the quality of the incorporated datasets, any errors, biases, or inconsistencies in these datasets may negatively impact the results generated by the system.

\bibliography{anthology,custom}
\bibliographystyle{acl_natbib}

\appendix

\section{Knowledge Verification}
According to Table~\ref{tab:hall_ratio}, our analysis reveals an increase in the overall hallucination rate when generating extended turn instruction data. To address this, \textsc{Mimir} incorporates a fine-grained knowledge verification feature for the generated datasets. Users can select any round of instruction data and verify it with a single-click action. For this purpose, we extract key QA pairs and topics and integrate them into our verification module. This module operates on a domain-specific state-of-the-art model. Utilizing this approach, we aim to generate more accurate and reliable responses. Currently, \textsc{Mimir} employs GPT-4 as its verification model, leveraging its exceptional performance across various medical tasks.
\begin{table}[ht]
    \centering
    \begin{tabular}{c|c}
    \toprule
         Turn & Overall Hallucination Ratio\\
    \midrule
         1 & 4.27\\
    \midrule
         2 & 7.37\\
    \midrule
         3 & 14.27\\
    \midrule
         4 & 21.27\\
    \midrule
         5 & 24.27\\

    \bottomrule
    \end{tabular}
    \caption{Halluciation Ratio across all instruction data generated from domain-specific datasets. Overall scores are reported by averaging all the results by domain-specific expert evaluation.}
    \label{tab:hall_ratio}
\end{table}

\section{Implements Details for Role-Playing}
\subsection{Memory Setting}
\begin{lstlisting}[caption= Memory setting for the running loop in \textsc{Mimir} agent system.]
for name in picked_roles:
    prompt = "You are {}. {} You come to a chat room 
    because you want to discuss the topic 
    about {}. " \
    "The following people are in 
    this chat room: {}. 
    What is your main point? Be brief, " \
    "and use at most 20 words
    and answer from your 
    perspective.".format(
        name, role_prompt[name], query,
        ', '.join(picked_roles))
    ideas[name] =generate(prompt_meta.format(prompt)
    , asure, ai_temperature)
   
\end{lstlisting}

In MIMIR, we present a framework designed to simulate interactive role-based dialogues in a chat room environment. Our methodology encompasses four key components: the initialization of a memory data structure for each role, the preparation of a compressed memory counterpart, the establishment of a placeholder for role-specific ideas, and the generation of these ideas through a sophisticated prompt formulation. By iterating over a predefined set of roles, our system dynamically constructs context-specific prompts, incorporating role-specific cues and a central discussion topic. This is followed by the generation of concise, perspective-driven responses using an advanced language model. 

\subsection{Memory Rater}
\begin{lstlisting}[caption= Memory setting for the running loop in \textsc{Mimir} agent system.]
import re
def get_rating(x):
    nums = [int(i) for i in re.findall(r'\d+', x)]
    if len(nums) > 0:
        return min(nums)
    else:
        return None

memory_ratings = {}
for name in picked_roles:
    memory_ratings[name] = []
    for i, memory in enumerate(memories[name]):
        prompt = "You are {}. Your ideas are: {}. 
        You are currently in a chat room and 
        you are talk about {}. " \
                 "You observe the following: {}. 
                 Give a rating, between 1 and 5, 
                 to how much you care about this. "
        .format(name, ideas[name], query, memory)
        res = generate(prompt_meta.format(prompt), 
        asure, ai_temperature)
        rating = get_rating(res)
        max_attempts = 2
        current_attempt = 0
        while rating is None and current_attempt 
        < max_attempts:
            rating = get_rating(res)
            current_attempt += 1
        if rating is None:
            rating = 0
        memory_ratings[name].append((res, rating))
   
\end{lstlisting}
In the given code, a function named get\textunderscore rating is implemented using regular expressions to extract numerical values from a string. The smallest number in the string is returned as the rating, or None if no numbers are found. The script iterates over a set of predefined roles stored in picked\textunderscore roles, managing each role's memories. For every memory associated with a role, a specific prompt is generated that includes the role's name, ideas, a query, and the memory itself. This prompt is then processed by an AI generation function.

The get\textunderscore rating function is subsequently used to analyze the AI's response, extracting a numerical rating that reflects the relevance or importance of the memory in question. In cases where no rating can be determined after a few attempts, a default rating of zero is assigned. The outcomes, comprising both the AI's response and the extracted rating, are stored in a dictionary, with each entry corresponding to a different role.

\section{Diologue Example}
\begin{table*}[ht]
    \centering %
    \begin{tabular}{|p{1cm}|p{3cm}|p{10cm}|} %
      \hline
      Round & Human & AI \\
      \hline
      1 & Recently, I've been having headaches and a sore throat. In the morning, I feel nauseous and have dry heaves when brushing my teeth. What should I do? & It is important to consult a healthcare professional if you are experiencing persistent headaches, sore throat, nausea, and dry heaves. These symptoms could be indicative of a variety of conditions, including a viral or bacterial infection, allergies, or even a more serious illness. Your healthcare provider can help determine the cause of your symptoms and recommend appropriate treatment. In the meantime, it may be helpful to rest, stay hydrated, and avoid triggers that may exacerbate your symptoms. \\
      \hline
      2 & What kind of triggers should I avoid? & The triggers you should avoid depend on the underlying cause of your symptoms. However, some general tips to help alleviate symptoms include:

        1. Avoiding exposure to allergens such as pollen, dust, and pet dander.
        
        2. Drinking plenty of fluids to stay hydrated.
        
        3. Avoiding caffeine, alcohol, and tobacco, which can irritate the throat and exacerbate nausea.
        
        4. Eating small, frequent meals throughout the day instead of large meals.
        
        5. Avoiding spicy, acidic, or fatty foods that can trigger nausea and acid reflux.
        
        6. Getting plenty of rest and avoiding strenuous activity until you feel better.
        
        It's important to consult with your healthcare provider to determine the specific triggers you should avoid based on your symptoms and medical history. \\
      \hline
      3 & Can stress be a trigger for these symptoms? & Yes, stress can be a trigger for headaches, sore throat, nausea, and dry heaves. Stress can cause physical symptoms such as tension headaches, muscle tension, and gastrointestinal distress. It can also weaken the immune system, making you more susceptible to infections that can cause sore throat and nausea. Additionally, stress can disrupt sleep, which can exacerbate symptoms such as headaches and nausea. If you suspect that stress may be contributing to your symptoms, it's important to take steps to manage your stress levels. This may include practicing relaxation techniques such as deep breathing, meditation, or yoga, getting regular exercise, and making time for activities that you enjoy. If you're having difficulty managing stress on your own, consider talking to a mental health professional who can help you develop coping strategies. \\
      \hline
    \end{tabular}
    \caption{Two agents are taking about the topic, "Recently, I've been having headaches and a sore throat. In the morning, I feel nauseous and have dry heaves when brushing my teeth. What should I do?"}
    \label{app:instrucion_example}
\end{table*}

\end{document}